\documentclass[10pt,twocolumn,letterpaper]{article}

\usepackage[pagenumbers]{cvpr} 

\usepackage{graphicx}
\usepackage{amsmath}
\usepackage{amssymb}
\usepackage{booktabs}
\usepackage{bbm}
\usepackage{multirow}
\usepackage{subcaption}
\usepackage{url}
\usepackage{paralist}

\makeatletter
\def\hlinew#1{%
	\noalign{\ifnum0=`}\fi\hrule \@height #1 \futurelet
	\reserved@a\@xhline}
\makeatother%
\usepackage[pagebackref=true,breaklinks=true,letterpaper=true,colorlinks,citecolor=blue,linkcolor=red,bookmarks=false]{hyperref}

\usepackage[capitalize]{cleveref}
\crefname{section}{Sec.}{Secs.}
\Crefname{section}{Section}{Sections}
\Crefname{table}{Table}{Tables}
\crefname{table}{Tab.}{Tabs.}
\def\cvprPaperID{3} 
\def\confName{CVPRW}
\def\confYear{2024}

\begin{document}

\title{Dual-Path Multi-Scale Transformer for High-Quality Image Deraining}

\author{Huiling Zhou \quad Xianhao Wu \quad Hongming Chen \\
College of Electronic Information Engineering, Shenyang Aerospace University
\\
}

\maketitle

\begin{abstract}
Despite the superiority of convolutional neural networks (CNNs) and Transformers in single-image rain removal, current multi-scale models still face significant challenges due to their reliance on single-scale feature pyramid patterns.
In this paper, we propose an effective rain removal method, the dual-path multi-scale Transformer (DPMformer) for high-quality image reconstruction by leveraging rich multi-scale information. This method consists of a backbone path and two branch paths from two different multi-scale approaches. Specifically, one path adopts the coarse-to-fine strategy, progressively downsampling the image to 1/2 and 1/4 scales, which helps capture fine-scale potential rain information fusion. Simultaneously, we employ the multi-patch stacked model (non-overlapping blocks of size 2 and 4) to enrich the feature information of the deep network in the other path. To learn a richer blend of features, the backbone path fully utilizes the multi-scale information to achieve high-quality rain removal image reconstruction. Extensive experiments on benchmark datasets demonstrate that our method achieves promising performance compared to other state-of-the-art methods.
\end{abstract}

\section{Introduction}
Due to the adverse effects of rain on human perception and computer vision, images are often affected by raindrops or rain streaks, resulting in blurriness and degradation.
Therefore, single image deraining techniques are of great significance in achieving high-quality restoration of the damaged images. 
However, addressing the limited availability of rainy images is a challenging problem. Traditional model-based methods\cite{kang2011automatic,chen2013generalized,li2016rain} typically rely on applying various mathematical and statistical priors to restore clearer images. 
Nevertheless, these manually designed priors exhibit low sensitivity to complex textures and fine details, leading to suboptimal restoration results.

Recently, numerous CNN-based frameworks\cite{fu2017clearing,li2018recurrent,yi2021structure,yang2020single,zamir2021multi} have been widely applied in the field of image deraining, achieving certain performance improvements through the convolutional capture of local features and weight sharing. 
However, CNN-based architectures are inherently limited by their local receptive fields, making it challenging to remove complex and long-range rain effects. Transformer-based methods\cite{wang2022uformer,zamir2022restormer,xiao2022image,chen2023learning,chen2024bidirectional} possess the modeling ability of long-range dependencies, effectively address the limitation. They can better capture global contextual information and demonstrate the significant restoration performance. 
Despite the advantages of Transformer methods in terms of global correlations, existing Transformer-based approaches mostly rely on single-scale feature pyramids \cite{li2017single,zhang2018density}. When restoring clear images, it lacks sufficient consideration for multi-scale features. 
Images often contain structures and features at multi-scale, such as fine textures and large-scale global structures, single-scale methods may not fully capture this diversity, leading to a performance decrease in deraining tasks.

\begin{figure}[!t]
	\centering 	
\includegraphics[width=\columnwidth]{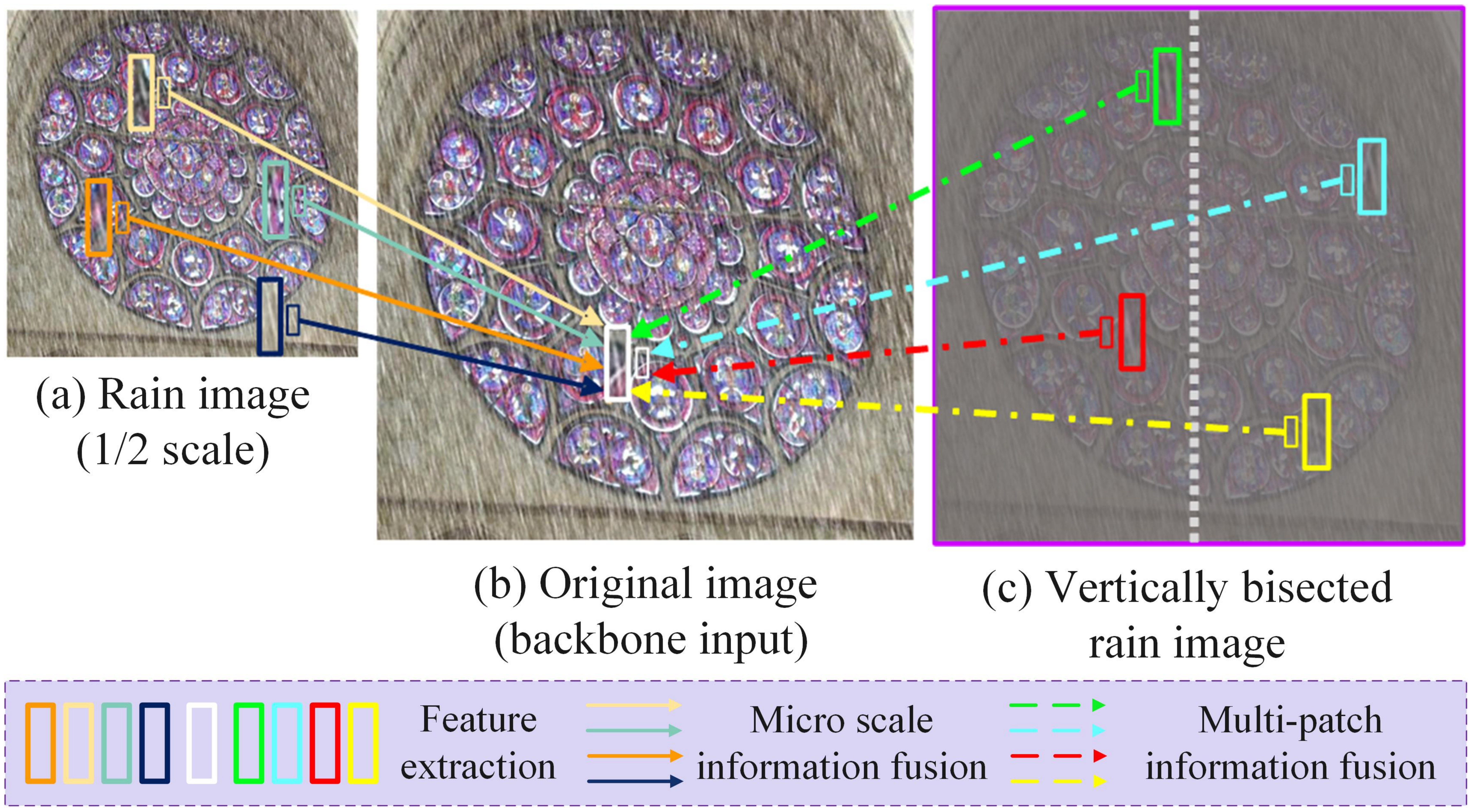}
    \caption{Propagation of dual-path rain streaks information flow.(a) represents the coarse-to-fine rain removal strategy adopted for half-scale images with rain; (c) is the multi-patch rain removal operation adopted for vertically bisected images with rain; (b) is the backbone input rain image imported by dual-path multi-scale image information flow. }
	\label{fig1}
 \end{figure}

We note that typical  multi-scale pyramid images \cite{fu2019lightweight,jiang2020multi} currently include two commonly used approaches: a) multi-patch ; b) coarse-to-fine , where images at different scales may carry complementary information to represent the target rain streaks.
The coarse-to-fine strategy is composed of multiple stacked subnetworks, each taking a downscaled image and progressively restoring it to a clear image in the manner. 
It explores multi-scale collaboration and complementary information of rain streaks in hierarchical deep features.
Multi-patch processes images with rain through a layered stacked representation, introducing additional patches to increase the depth of information for model features. 
Despite adopting multiscale methods, these strategies are still limited by the single-directional information flow, leading to inaccurate estimates of small-scale image features and significant information propagation errors in the process of high-quality image restoration.

Towards this goal, we propose a dual-path multi-scale Transformer for high-quality image deraining. Specifically, our study aims to integrate multi-scale features more comprehensively through different information propagation paths.
On one path, it receives fine-scale potential feature information from the coarse-to-fine process.
On the other path, it receives stacked block information of the same scale but different patch divisions from the multi-patch approach.
With the dual-path multi-scale method, the network can simultaneously acquire and fuse information from different directions, thereby better capturing the rich expression of multi-scale characteristics in the image.

The main contributions are summarized as follows:
\begin{compactitem}
	\item We propose a dual-path multi-scale Transformer network model.

	\item We integrate the coarse-to-fine strategy and the multi-patch operation to effectively utilize the rain information in the input image, enriching inter-layer feature transformations to achieve high-quality deraining outputs.

	\item Extensive experimental results demonstrate that our dual-path multi-scale Transformer network outperforms existing state-of-the-art (SOTA) approaches on commonly used benchmark datasets.
\end{compactitem}

\section{Related Work}
In this section, we provide a concise overview of the recent advancements in single-image deraining and multi-scale methods, respectively.

\subsection{Single Image Deraining}
Current deep-learning based approaches \cite{chen2022unpaired,zamir2022restormer} have achieved remarkable results in image rain streaks removal, especially in progressive restoration performance on numerous CNN-based frameworks \cite{fu2017clearing,li2018recurrent,yi2021structure,yang2020single,zamir2021multi}. Zhang et al.\cite{zhang2018density} propose a density-aware multi-stream based on dense connected CNN algorithm, DID-MDN, effectively removed the corresponding rain streaks with density estimation. Later, Yang et al.\cite{yang2017deep} constructe a cyclic expansion network of multi-task learning for the gradual removal of rain streaks by combining context information. By exactly learning the motion fuzzy kernel and ResNet architecture, Wang et al. \cite{wang2020rain} propose an area line framework (KGCNN) for the kernel guided CNN to resolve motion blur bug generated by rain line.
To model remote dependencies for non-local information capture, there is a tendency for Transformer to be widely applied in the field of image restoration \cite{wang2022uformer,zamir2022restormer,xiao2022image,chen2023learning} and perform significantly better than the previous CNN baselines. Wang et al.\cite{wang2022uformer} utilizes a  locally-enhanced window Transformer block and a learnable multi-scale restoration modulator to achieve better results. To better capture distant pixel interactions, Zamir et al.\cite{zamir2022restormer} introduce an efficient Transformer model, Restormer, overcoming the issue of quadratic complexity limitation. Additionally, Xiao et al. \cite{xiao2022image} propose an efficient Transformer-based architecture for image deraining (IDT) , employing general visual task priors and complementary transformers to capture both local and distant features. Despite significant advancements of CNNs in the field of image deraining, they still exhibit certain limitations in handling long-range dependencies and global relationships. In this work, we opt for Transformer as our network backbone to achieve superior restoration performance and flexibility.

\begin{figure*}[t]
	\centering
	\includegraphics[width=1.0\textwidth]{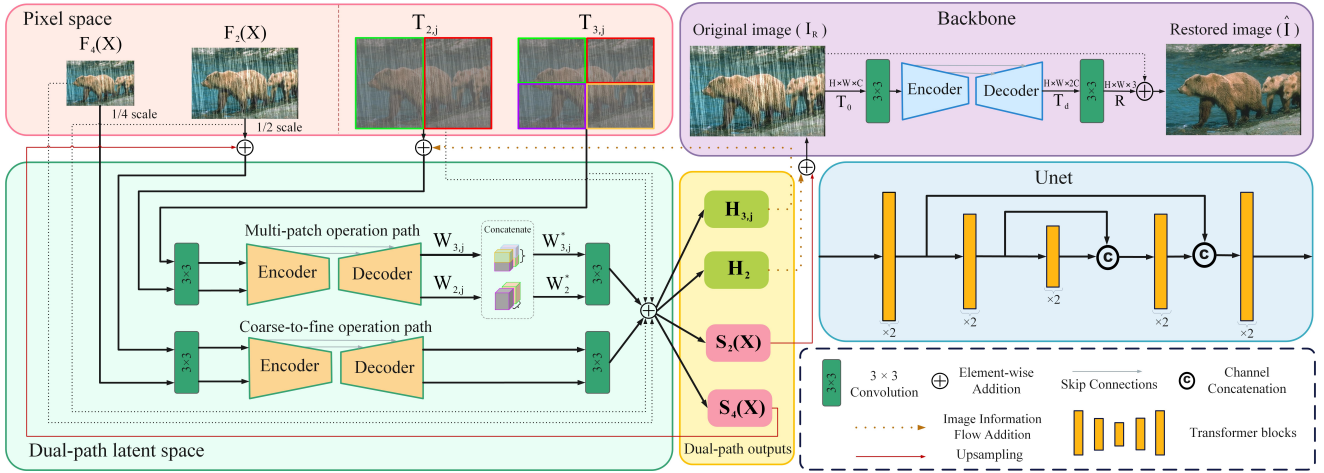}
	\caption{The overall architecture of dual-path multi-scale Transformer (DPMformer) proposed in this paper mainly includes image pixel space, backbone network, multi-patch overlay layer network, coarse-to-fine strategy path and an encoding and decoding structure (UNet) composed of multiple Transformer blocks.}
	\label{fig2}
\end{figure*}
\subsection{Multi-scale Learning}
Rain streaks in the air are influenced by a myriad of factors, particularly exhibiting certain shared characteristics at various scales. 
Consequently, it is imperative to investigate methods for effectively integrating rain line information across different scales to enhance image restoration. 
Nevertheless, the possible linkages among rain streaks at various sizes are not sufficiently addressed by most current deraining methods\cite{li2017single,zhang2018density,li2016rain} . 
Only a minority \cite{fu2017removing,liu2019removing} have focused on leveraging multiscale knowledge.
Fu et al. \cite{fu2019lightweight} present a lightweight pyramid network (LPNet) for single picture training that makes advantage of pyramid layer by layer breakdown to streamline the learning of rain streaks features.
With a residual attention mechanism, Zheng et al. provide a unique multiscale de-raining network that efficiently functions in a coarse-to-fine way via a pyramid architecture. 
In addressing the ill-posed inverse problem of single image deraining, Yang et al.\cite{yang2019single} present ReMAEN, a progressive method utilizing the Recurrent Multi-scale Aggregation and Enhancement Network.  

Distinguished by its symmetric structure and shared channel attention, it collaboratively selects useful information and removes rain streaks in a stage-wise manner. For better single-image deraining, Jiang et al.\cite{jiang2020multi} develope the Multi-Scale Progressive Fusion Network (MSPFN), which builds a multi-scale pyramid structure and recursively computes global texture capture, combining data from multiple scales to improve image restoration performance. 
By extracting and utilizing information from various scales, the multi-scale approach\cite{jiang2020multi} minimizes the impact on areas not affected by rain, thus reducing the creation of artifacts and maintaining the authenticity of the original scene.
From a broad to a detailed perspective, this method\cite{cho2021rethinking,fu2019lightweight} facilitates the reconstruction of images that more closely resemble reality, endowed with rich details and textures.

\section{Proposed Method}
In this section, Figure \ref{fig2} initially presents our dual-path overall structure. After discussing the main network, the two components of the multi-scale modules are specifically researched. 
We elaborate on the details of the network and provide a comprehensive overview of the loss functions in the following.
%

\subsection{Overall Pipeline}
Our network architecture consists of a backbone path and two branches from the multi-scale path. 
Given a rainy image $I_{R}\in\mathbb{R}^{H\times W\times3}$, in the backbone, the output image from the dual-path branches is first added to enhance details.
This combined output then passes through a standard $3\times3$ convolutional layer to obtain a low-level feature embedding $\mathbf{T}_{0}\in\mathbb{R}^{H\times W\times C}$ as the model input, where $H \times W$ represents the spatial resolution of the feature map and $\color{black}{C}$ denotes the number of channels. 
Subsequently, the shallow-level feature map extracted is converted into a deep-level feature image $\mathbf{T}_{d}\in\mathbb{R}^{H\times W\times2C}$ through a symmetric 3-layer encoder-decoder, ensuring the comprehensive extraction of rain streaks feature information.
The process concludes with a $3\times3$ convolution operation, when combined with the residual, produces clear outputs for the main backbone channel.
The two branches from the backbone path are described below.

{\flushleft\textbf{The Multi-patch operation path}.}
In combination with the backbone, it indicates the notation $N_{i \in[1,2,4]}$ to represent the increasing number of partitioned patches for the input original image at each level. Specifically, vertical partitioning at the second level and 4 partition at the third level.
We represent the original input rainy image as $\mathbf{T}_{1}$, where $\mathbf{T}_{i,j}$ is the $j$-th image block in the $i$-th layer. 
Additionally, $E_{i}$ and $D_{i}$ denote the encoder and decoder of the $i$-th layer, $\mathbf{W}_{ij}$ is the output of $\mathbf{T}_{ij}$ after passing through $E_{i}$ and $D_{i}$, and $\mathbf{H}_{ij}$ is the output image block at each layer. Each layer of the network architecture consists of an encoder-decoder pair stacked with transformer blocks. 
The inputs $N_{i \in[1,2,4]}$ are obtained by segmenting the original input image $\mathbf{T}_{1}$ into non-overlapping blocks.
The outputs of the encoder and decoder at finer grids are progressively added to the network of the previous level to incorporate all feature information for finer-level inference.
The multi-patch deraining \cite{zhang2019deep}  process from the dual-path starts at the bottom level 3.
$\mathbf{T}_{1}$ is divided into 4 non-overlapping blocks $\mathbf{T}_{3,j},j=1,\cdots,4$.These blocks are input to $E_{3}$ and $D_{3}$ after passing through a $3\times3$ convolution, resulting in the following feature representation:
\begin{equation}
 \mathbf{W}_{3,j}=D_3[E_3(\mathbf{Conv}_{3*3}(\mathbf{T}_{3,j}))],j\in\{1\ldots4\}.   
 \label{eq:patch level-3 edout}
\end{equation}

Afterwards, we concatenate the adjacent feature blocks to obtain an entirely new feature representation $W_{3,j}^{*}$, which has the same size as the input feature blocks at level 2:
\begin{equation}
\mathbf{W}_{3,j}^*=\mathbf{W}_{3,2j-1}\textcircled{c}\mathbf{W}_{3,2j},j\in\{1,2\},
\label{eq:patch level-3 cat}
\end{equation}
where $\textcircled{c} $ denotes the concatenation operator. 
The concatenated feature representation $W_{3,j}^{*}$ is then added to the input image of level 2 after a $3\times3$ convolution to obtain the outputs of the level 3:
\begin{equation}
\mathbf{H}_{3,j}=\mathbf{Conv}_{3*3}(\mathbf{W}_{3,j}^*)+\mathbf{T}_{3,j},j\in\{1\ldots4\}.  
\label{eq:patch level-3 out}
\end{equation}

Next, we move up to the level 2, where the input consists of vertically halved images $\mathbf{T}_{2,j},j=1,2.$ After a $3\times3$ convolution and concatenation with the feature images from the third level $W_{3,j}^*$, we obtain $\mathbf{T}_{2,j}^{*}$, which is then input into $E_{2}$ and $D_{2}$:
\begin{equation}
\mathbf{W}_{2,j}=D_3[E_3(\mathbf{T}_{2,j}^*)],j\in\{1,2\},
\label{eq:patch level-2 edout}
\end{equation}
\begin{equation}
\mathbf{W}_2^*=\mathbf{W}_{2,1}\textcircled{c}\mathbf{W}_{2,2}.
\label{eq:patch level-2 cat}
\end{equation}

Similarly, after passing through a $3\times3$ convolution and combining with the residual from the degraded image, we obtain the output image at level 2:
\begin{equation}
\mathbf{H}_2=\mathbf{Conv}_{3*3}(\mathbf{W}_2^*)+\mathbf{T}_{2,j},j\in\{1,2\}.
\label{eq:patch level-2 out}
\end{equation}

The following describes the network model of the other branch.

{\flushleft\textbf{The coarse-to-fine operation path}.}
For the given primary raw input rainy image T1, it first utilizes a Gaussian kernel to create a Gaussian pyramid of the rainy images, downsampling T1 to various scales.
Starting with the image downscaled to 1/2 scale and 1/4 scale\cite{jiang2020multi} as inputs, the process begins with the 1/4 scale image undergoing a $3\times3$ convolution to extract shallow features,followed by processing through the same encoder-decoder architecture as used in the previous branch. Finally, through a $3\times3$ convolution and residual connection with the degraded image, a 1/4 scale smaller-scale image output is obtained,which is subsequently upsampled and added to the 1/2 scale input image:
\begin{equation}
\mathbf{F}_2^*(\mathbf{X})=F_2(\mathbf{X})+U(S_4(\mathbf{X})),
\label{eq:coarse level-3 up to 2}
\end{equation}
where $F_2(\mathbf{X})$ represents the 1/2 scale input from the coarse-to-fine network levels, $S_4(\mathbf{X})$ is the output image at 1/4 scale,and $U()$ stands for upsampling process. The function $\mathbf{F}_2^*(\mathbf{X})$ denotes the input representation at 1/2 scale, obtained by upsampling the coarser network (the previous level) using a Gaussian kernel and adding it to the original scale input. 
Introducing feature information from rough images (small scale) into the detailed pictures (large scale) is advantageous for extracting the clear feature information potentially hidden within the small scale.
Ultimately, this approach channels effective information flow into the main input network, where the core part integrates the rich features from both paths, significantly enhancing the main output image's restoration performance.

\subsection{UNet Architecture}
In our dual-path multi-scale network, each level comprises an encoder and a decoder, forming a symmetric 3-level U-Net structure.
Each encoder-decoder pair at every level contains multiple transformer blocks, with a consistent number of blocks set at [2,2,2].
Starting from the $H\times W$ spatial input, the encoder progressively reduces the resolution while extending the channel capacity.
The decoder takes $\mathbf{T}_l\in\mathbb{R}^{\frac H4\times\frac W4\times4C}$ as input to restore high-resolution displays. 
To enhance the completeness of the restoration process, we establish skip connections between the encoder features and decoder, facilitating the thorough aggregation of low-level and high-level image features.
 After each concatenation operation, a $1\times1$ convolution is applied to halve the channel numbers. For upsampling and downsampling, we employ pixel-shuffle and pixel-unshuffle techniques, respectively.
Finally, convolutional layers are used to process the refined features, generating a residual image $\mathbf{R}$.
This residual image is added to the original image to obtain the restored image  $\mathbf{\hat{I}}=I_{R}+\mathbf{R}$. Next, we will introduce the MDTA and GDFN modules \cite{zamir2022restormer} of the transformer.

For traditional self-attention mechanisms, the spatio-temporal complexity of key-query dot product interactions grows quadratically with the input spatial resolution, making it challenging to achieve high-resolution image restoration tasks due to its high complexity.
Therefore, we utilize MDTA, which incorporates depth-wise convolution and computes cross-covariance across channels, to generate global-local contextual attention maps.
The overall computation process of MDTA is defined as follows:
\begin{equation}
\text{MDTA}\left(\hat{\mathbf{Q}},\hat{\mathbf{K}},\hat{\mathbf{V}}\right)=\hat{\mathbf{V}}\cdot\text{Softmax}\left(\hat{\mathbf{K}}\cdot\hat{\mathbf{Q}}/\alpha\right),
\label{eq:atten}
\end{equation}
where $\alpha $ is a learnable scaling parameter before using the softmax function.
Similarly, we partition the channel dimension into multiple parallel and independent heads to learn attention maps.

The feedforward neural network plays a crucial role in providing the required nonlinear processing for the Transformer model and its variants.
In this work, we introduce GDFN, which specifically aims to enhance the flow of fine-detail contextual information at each layer, selectively allowing/blocking the flow of information through a gating mechanism where one path in two parallel linear transformation layers is nonlinearly activated by GELU. 
Like MSFN\cite{chen2023learning} , GDFN also leverages deep convolutions to improve multi-scale local information processing, expand the effective receptive field, and further enhance the detailed feature processing for enhancing the image's rain removal capability.

\begin{table}[t]
	\centering
	\caption{Comparison of individual and average quantitative results based on the Rain200L/H dataset. \textbf{Bold} and \underline{underline} indicate the best and second-best results.
P stands for prior-based methods, C stands for CNN-based methods, and T stands for Transformer-based methods.}
	\resizebox{1.0\columnwidth}{!}{
	\begin{tabular}{cc|cc|cc|cc}
	\hlinew{1.0pt}
	\multicolumn{2}{c|}{Datasets}                                                   & \multicolumn{2}{c|}{\textit{Rain200L}}    & \multicolumn{2}{c|}{\textit{Rain200H}}    & \multicolumn{2}{c}{\textit{Average}}  \\ \hline
	\multicolumn{2}{c|}{Metrics}                                                    & PSNR           & SSIM            & PSNR           & SSIM            & PSNR           & SSIM                     \\ \hline
	\multicolumn{1}{c|}{\multirow{2}{*}{P}}       & DSC \cite{luo2015removing}          & 27.16          & 0.8663          & 14.73          & 0.3815          & 20.95          & 0.6239               \\
	\multicolumn{1}{c|}{}                                           & GMM \cite{li2016rain}          & 28.66          & 0.8652          & 14.50          & 0.4164          & 21.58          & 0.6408                  \\ \hline
	\multicolumn{1}{c|}{\multirow{8}{*}{C}}         & DDN \cite{fu2017removing}          & 34.68          & 0.9671          & 26.05          & 0.8056          & 24.87          & 0.8864                 \\
	\multicolumn{1}{c|}{}                                           & RESCAN \cite{li2018recurrent}       & 36.09          & 0.9697          & 26.75          & 0.8353          & 31.42          & 0.9025                 \\
	\multicolumn{1}{c|}{}                                           & PReNet \cite{ren2019progressive}       & 37.80          & 0.9814          & 29.04          & 0.8991          & 33.42          & 0.9403                  \\
	\multicolumn{1}{c|}{}                                           & MSPFN \cite{jiang2020multi}       & 38.58          & 0.9827          & 29.36          & 0.9034          & 33.97          & 0.9431          \\
	\multicolumn{1}{c|}{}                                           & RCDNet \cite{wang2020model}       & 39.17          & 0.9885          & 30.24          & 0.9048          & 34.71          & 0.9467               \\
	\multicolumn{1}{c|}{}                                           & MPRNet \cite{zamir2021multi}      & 39.47          & 0.9825          & 30.67          & 0.9110          & 35.07          & 0.9468                     \\
	\multicolumn{1}{c|}{}                                           & DualGCN \cite{fu2021rain}       & 40.73          & 0.9886          & 31.15          & 0.9125          & 35.94          & 0.9506             \\
	\multicolumn{1}{c|}{}                                           & SPDNet \cite{yi2021structure}       & 40.50          & 0.9875          & 31.28          & 0.9207          & 35.89          & 0.9541                  \\ \hline
	\multicolumn{1}{c|}{\multirow{5}{*}{T}} & Uformer \cite{wang2022uformer}    & 40.20          & 0.9860          & 30.80          & 0.9105          & 35.50          & 0.9483                \\
	\multicolumn{1}{c|}{}                                           & Restormer \cite{zamir2022restormer}    & 40.99         & 0.9890         & 32.00          & \underline{0.9329}          & 36.50          & 0.9610           \\
	\multicolumn{1}{c|}{}                                           & IDT \cite{xiao2022image}          & 40.74          & 0.9884          & 32.10        & \textbf{0.9344}          & 36.42          & \underline{0.9614}            \\
	\multicolumn{1}{c|}{}                                           & {DRSformer} \cite{chen2023learning} & \underline{41.23} &\underline{0.9894} & \textbf{32.17} & 0.9326 & \underline{36.70} & 0.9610  \\ 
         \multicolumn{1}{c|}{}                                          & \textbf{Ours} & \textbf{41.81} & \textbf{0.9905} & \underline{32.14} & \underline{0.9329} & \textbf{36.98} & \textbf{0.9617} \\ \hlinew{1.0pt}
\end{tabular}
}	
   \label{table1}	
\end{table}
\subsection{Loss Function}
Due to the unavailability of precise ground truth, developing an effective loss function is essential to regularize network training for better fitting.
In this study, we employ the Charbonnier penalty function to reduce the error with respect to the actual rain streaks distribution,the function is represented as:
\begin{equation}
L_{char}=\sqrt{(I_{R}^{*}-I_{R})^{2}+\varepsilon^{2}}.
\label{eq:loss char}
\end{equation}

In Equation \ref{eq:loss char}, $I_{R}^{*}$ represents the predicted residual rain image.
The predicted rain-free image $I_{derain}$ is obtained by subtracting $I_{R}^{*}$ from the rainy image $I_{R}$. The smoothing coefficient $\varepsilon$ is set to $10^{-3}$.
To ensure the fidelity of high-frequency texture information during rain streaks removal, we propose an edge loss to constrain the high-frequency information between the ground truth $I_{clean}$ and the predicted rain-free image $I_{derain}$.
The edge loss is defined as: 
\begin{equation}
L_{edge}=\sqrt{(lap(I_{clean})-lap(I_{derain}))^2+\varepsilon^2}.
\label{eq:loss edge}
\end{equation}

In Equation \ref{eq:loss edge}, $I_{clean}$ and $I_{derain}$ are edge maps extracted using the Laplacian operator. Subsequently, guided by the fftLoss, we quantify the average discrepancy in the frequency domain between the ground truth and the predicted rain streak image as follows:
\begin{equation}
L_{fft}=\sum_{k=1}^K\frac1{t_k}\parallel\mathcal{F}(I_{R_k}^*)-\mathcal{F}(I_{R_k})\parallel_1.
\label{eq:loss fft}
\end{equation}

Based on the considerations above, the final loss function is determined as:
\begin{equation}
L_{total}=\lambda_1L_{char}+\lambda_2L_{edge}+\lambda_3L_{fft},
\label{eq:loss total}
\end{equation}
where $\lambda_{i}$ represents the dynamic weight values. In our experiments, we empirically set $\lambda_{1}=1$, $\lambda_{2}=0.05$, $\mathrm{and}$ $\lambda_{3}=0.01$.

\section{Experiments}
In this section, to evaluate the effectiveness of our proposed dual-path multi-scale approach, we compare our method's performance with 14 state-of-the-art rain removal methods.
These representative methods include two prior-based models (DSC\cite{luo2015removing} and GMM\cite{li2016rain}), eight CNN-based methods (DDN\cite{fu2017removing}, RESCAN\cite{li2018recurrent}, PReNet\cite{ren2019progressive}, MSPFN\cite{jiang2020multi}, RCDNet\cite{wang2020model}, MPRNet\cite{zamir2021multi}, DualGCN\cite{fu2021rain}, SPDNet\cite{yi2021structure}), and the recent four Transformer-based methods (Uformer\cite{wang2022uformer}, Restromer\cite{zamir2022restormer}, IDT\cite{xiao2022image}, and DRSformer\cite{chen2023learning}).

\begin{figure}[!t]
	\centering 	
\includegraphics[width=\columnwidth]{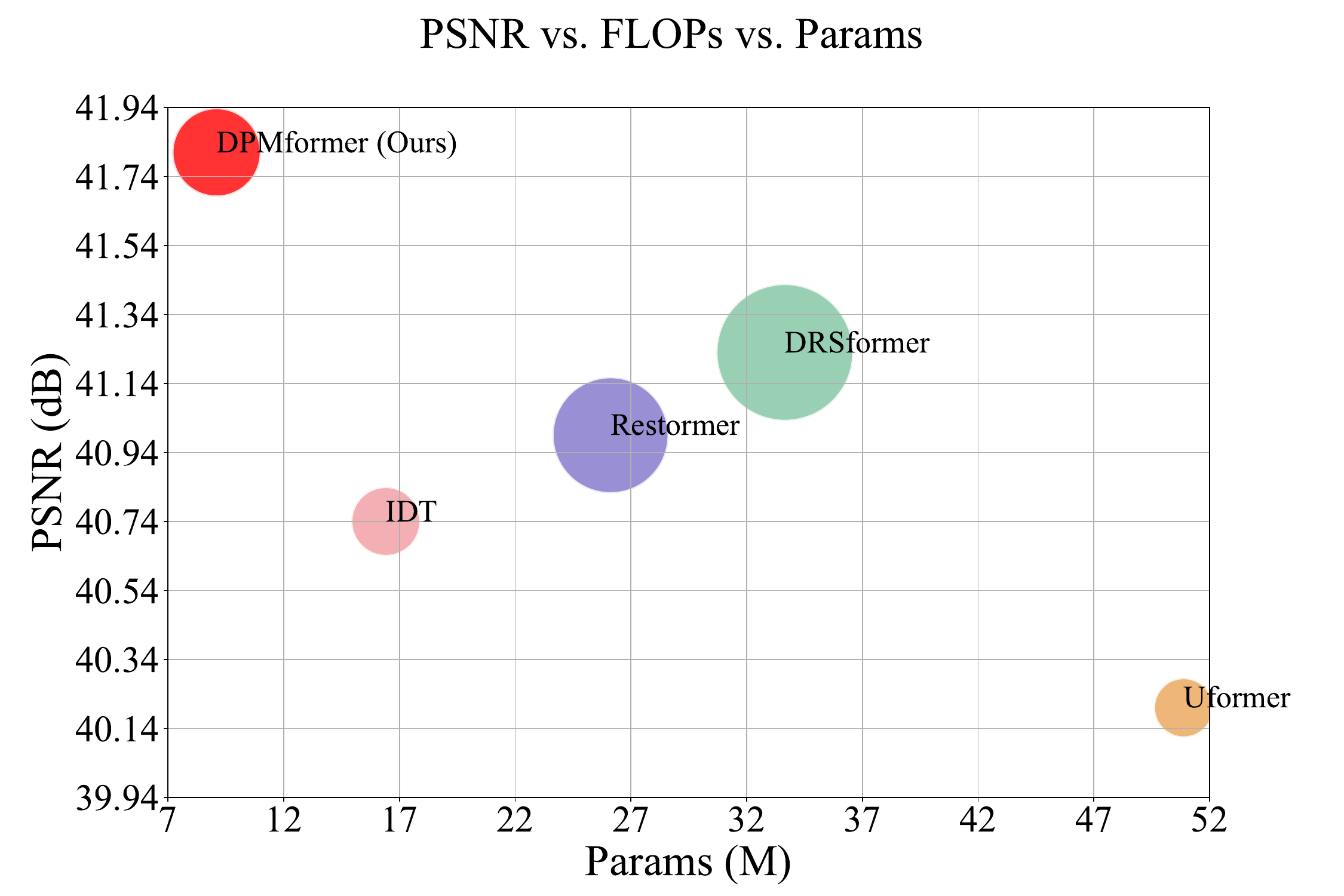}
    \caption{Model complexity and performance comparisons of the proposed method and other state-of-the-art models on the Rain200L dataset in terms of PSNR, model parameters and FLOPs. The area of each circle denotes the number of FLOPs. Here, FLOP calculation is based on image sizes of $256 \times 256$.}
	\label{fig3}
	 \end{figure}
 
\subsection{ Datasets and Metrics}
We trained/evaluated our method on datasets: Rain200L and  Rain200H. 
The Rain200L/H datasets both consist of 1,800 synthetic rainy images for training and 200 for testing each. 
Following the previous rain removal methods, we use Peak Signal-to-Noise Ratio (PSNR)\cite{cho2021rethinking} and Structural SIMilarity (SSIM)\cite{zhang2019deep} as quantitative evaluation metrics for the benchmarks mentioned above.Note that,We calculate the values of PSNR and SSIM using the Y channel \cite{chen2023learning,li2023dilated} in the YCbCr space.

\begin{figure*}[t]
	\centering
	\includegraphics[width=1.0\textwidth]{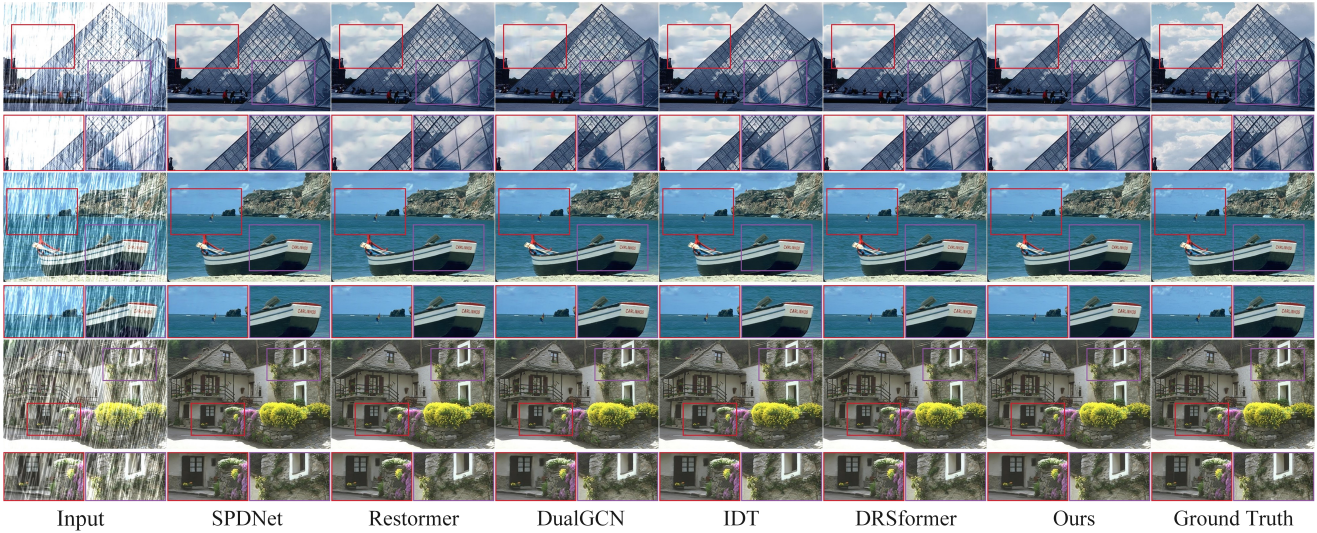}
	\caption{Visual quality comparison on the Rain200H dataset. Zooming in the figures offers a better view at the deraining capability.}
	\label{fig4}
	\end{figure*}

\subsection{ Implementation Details}
In our Unet, the three-level symmetric encoder-decoder consists of Transformer blocks with a count of [2, 2, 2] while 
the number of attention heads in MDTA is set as [1, 2, 4]. The initial channel number C is 48, and the expansion factor parameter in GDFN is 2.66. We utilize the Adam optimizer (with parameters $\beta_1=0.9, \beta_2=0.999$, weight decay $1\times10^{-8}$).
We train for 500 epochs on Rain200L/H, 200 epochs on DID-Data and DDN-Data, and 10 epochs on SPA-Data. 
The initial learning rate is fixed at $1\times10^{-4}$ and gradually reduced to $1\times10^{-6}$ using cosine annealing. 
The entire framework is trained on PyTorch with 4 TESLA
V100 GPUs, setting a patch size of $256\times256$.

\begin{table}[t]\small
	\centering
	\caption{Ablation study for different variations on 250 epoches of our dual-path multi-scale transformer on the Rain200H dataset. Multi-patch and Coarse-to-fine denote two ways in multi-scale methods.}
	\resizebox{1.0\columnwidth}{!}{
	\begin{tabular}{cccccc}
	\hlinew{1.0pt}
	Models &  Multi-patch & Coarse-to-fine & PSNR   &  SSIM   \\ \hline
	(a)    & $\checkmark$       &           & 32.07      & 0.9314 \\
	(b)    &      & $\checkmark$         & 32.04         & 0.9304 \\
	(c)    & $\checkmark$      & $\checkmark$    & \textbf{32.11}   &\textbf{0.9320} \\ \hlinew{1.0pt}
    \end{tabular}
	}	
	\label{table2}
\end{table}

\subsection{Comparisons with the state-of-the-arts}
The quantitative evaluation results on commonly used synthetic benchmark datasets, as shown in Table \ref{table1}. It demonstrate that our method achieves the highest values in terms of PSNR and SSIM compared to state-of-the-art methods.
Particularly, our method outperforms the average performance of other Transformer-based methods in parallel by 0.70 dB on Average.
Visual comparison results on the Rain200H dataset is provided in Figure \ref{fig4}.

Figure \ref{fig3} clearly shows our DPMformer method under the comprehensive qualitative analysis of PSNR, Floating-point Operations (FLOPs) and parameters (Params) based on Rain200L dataset. Significantly better than the four Transformer-based approaches: Uformer \cite{wang2022uformer} , Restormer \cite{zamir2022restormer} , IDT \cite{xiao2022image} , and DRSformer \cite{chen2023learning}. Additionally, the number of parameters of our DPMformer method only reaches 9.09M, which is on average smaller than the DRSformer method 33.65M, realizing a more lightweight network architecture.

\begin{figure}[!t]
	\centering 	
\includegraphics[width=\columnwidth]{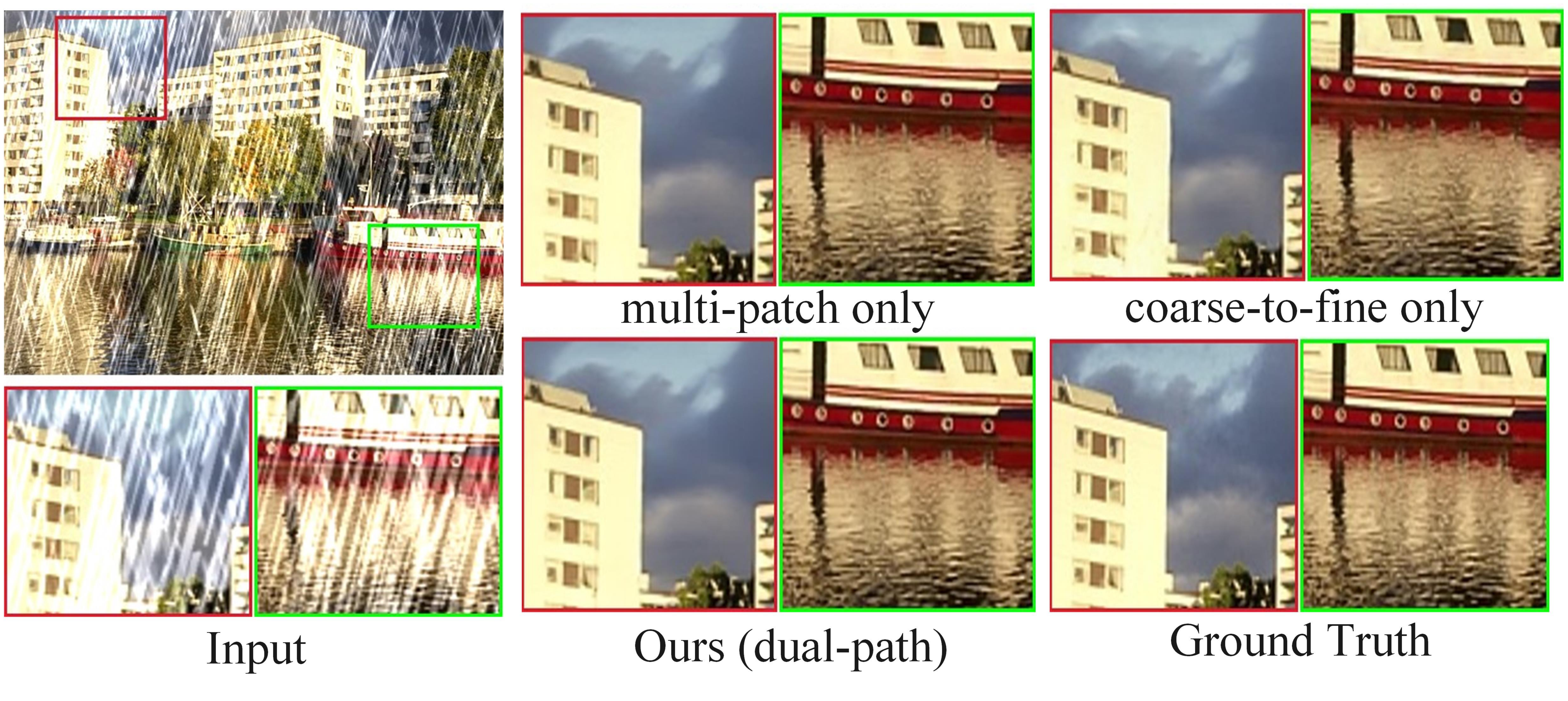}
    \caption{Qualitative analysis of ablation selection strategies.}
	\label{fig5}
 \end{figure}

\subsection{Ablation Studies}
To demonstrate the progressiveness of our framework, we investigate different variants of the proposed DPMformer. 
We mainly consider the following variants: (1) single-path multi-patch; (2) single-path coarse-to-fine. 
The 250 epoches of quantitative results based on Rain200H are presented in Table \ref{table2}.
It can be observed that our model (c) performs better than other possible configurations, indicating that the dual-path multi-scale strategy we designed provides a corresponding gain in the final performance of our DPMformer.
As shown in Figure \ref{fig5}, our method (c) has superior recovery performance.

\section{Conclusion}
In this paper, we propose a dual-path multi-scale Transformer (DPMformer) network structure to effectively address the image deraining problem.
We introduce the multi-patch and coarse-to-fine joint deraining network that combines small-scale hidden features with fine details from multi-block images, enhancing the performance across multiple scales.
At each encoding/decoding stage, we employ a Transformer-based Unet structure to facilitate precise integration of global and local detailed features. Experimental results demonstrate that our dual-path multi-scale Transformer exhibits superior deraining performance compared to state-of-the-art methods.

{\small
\bibliographystyle{ieee_fullname}
\bibliography{egbib}
}

\end{document}